\documentclass{article}

\usepackage[nonatbib,final]{nips_2017}
\usepackage[numbers]{natbib}

\usepackage[utf8]{inputenc} 
\usepackage[T1]{fontenc}    
\usepackage{hyperref}       
\usepackage{url}            
\usepackage{booktabs}       
\usepackage{amsfonts}       
\usepackage{microtype}      
\usepackage{graphicx}

\title{An interpretable latent variable model for attribute applicability in the Amazon catalogue}

\author{
  Tammo Rukat\thanks{Work was done at Amazon Berlin}\\
  Department of Statistics\\ University of Oxford\\
  \texttt{tammo.rukat@stats.ox.ac.uk}
  \And
  Dustin Lange\\
  Amazon\\ Berlin, Germany\\
  \texttt{cedrica@amazon.com}  
  \And
  C\'edric Archambeau\\
  Amazon\\ Berlin, Germany\\
  \texttt{langed@amazon.com}  
}

\begin{document}

\maketitle

\begin{abstract}
  Learning attribute applicability of products in the Amazon catalog (e.g., predicting that a \texttt{shoe}
  should have a value for \texttt{size}, but not for
  \texttt{battery-type}) at scale is a challenge. The need for an interpretable model is contingent on (1) the lack of ground truth training data, (2) the need to utilise prior information about the underlying latent space and (3) the ability to understand the quality of predictions on new, unseen data.
  To this end, we develop the MaxMachine, a probabilistic latent
  variable model that learns distributed binary representations,
  associated to sets of features that are likely to co-occur in the
  data. Layers of MaxMachines can be stacked such that higher layers
  encode more abstract information. Any set of variables can be
  clamped to encode prior information. We develop fast sampling based posterior inference.
  Preliminary results show that the model improves over the baseline in 17 out of 19 product groups and provides qualitatively reasonable predictions.
\end{abstract}

\section{Attribute Applicability}
\label{sec:org75c4eff}
\begin{figure}[h]
  \centering
\includegraphics[width=.65\textwidth]{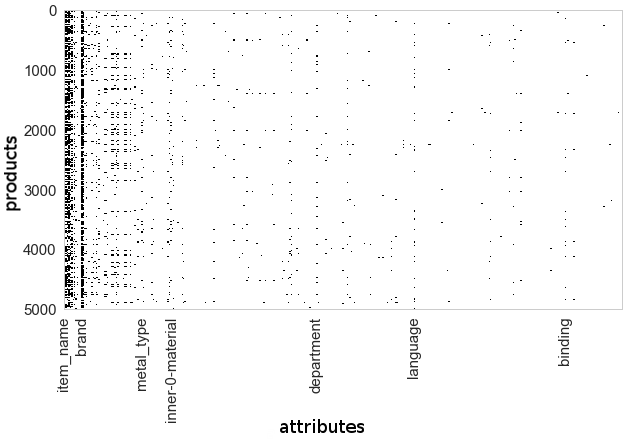}
\caption{
  Attribute applicability as binary matrix for a random subset of the German marketplace. Attributes are filtered for being applicable to at least 1\% of the sampled products. Black dots denote ones, indicating applied attributes. White dots denote zeros, indicating the absence of any attribute value.
  \label{fig:data}}
\end{figure}

Many real-world datasets can be viewed as object-by-attribute matrices. A prominent example is the Amazon catalogue which contains over 100 million products (objects) and hundreds of attributes, of which only a small subset is assigned to each product. Thus, product-attribute-assignment can be viewed as a sparse binary matrix, shown for a small subsample of the German Amazon marketplace in Fig.\space{}\ref{fig:data}.
Being able to distinguish between attributes that are truly
non-applicable (e.g., \texttt{battery-type} for a shoe), attributes that 
could reasonably be applied (e.g., \texttt{weight} for a book), and
attributes that are clearly applicable (e.g., \texttt{size} for a T-shirt) is
crucial for applications such as attribute imputation models, data
quality management, template generation, product comparison and
virtually all customer-facing downstream applications.

We can cast the task of predicting attribute applicability as a
multi-label classification problem, 
where each attribute constitutes a label and an arbitrary number 
of labels is assigned to each product. While there is recent progress
in such extreme multi-label classification problems
\cite{Bhatia2015,Jain2017}, we face a particular challenge:
The absence of reliable training labels makes it difficult to
define a training metric. Therefore, we approach attribute
applicability as an unsupervised problem and develop a probabilistic
latent variable model
that describes the generative process by which the binary
product/applicability matrix is generated from a set of latent
features. We aim to retain a simple,
interpretable model, resembling the process of a marketplace
seller who is filling in attributes for their product.
The rationale behind the model is that each latent
feature corresponds to a set of attributes that are likely to appear
together such as (\texttt{title}, \texttt{pages}, \texttt{language}, \texttt{release date}) or
(\texttt{width}, \texttt{height}, \texttt{length}).
Each of these sets is represented by a latent dimension and the
generative process for any product-attribute-pair is a noisy
disjunction of these feature sets. 
The model design is further motivated by keeping sampling-based
posterior inference scalable.
We also assume that additional product or attribute specific prior
information is available. Here, this is exemplified by the
\texttt{product type}, a string attribute that is assigned to a majority of products. Product types are curated such that there exist only
around 500 distinct values. They provide useful information that we need
to leverage for optimal performance. Moreover, attribute assignments
have a strong product type specific pattern which lets us use the
product type specific attribute frequency as a baseline probability estimate.

\section{MaxMachine Model}
\label{sec:orga18e9af}

A first candidate model is Boolean Matrix 
Factorisation \cite{Rukat2017},
where, both, the features and their allocations are
binary. The data generating process is a noisy matrix product between these
binary matrices whose result is thresholded at 1. With \(N\) products and \(D\)
attributes \(x_{nd} \in \{0,1\}\) denotes the application status for
a product-attribute-pair. We have $L$ latent dimensions and the factor matrices
are denoted as \(U \in \{0,1\}^{L\times D}\) and \(Z \in
  \{0,1\}^{N\times L}\), such that \(u_{l,d=1,\ldots, D}\) encodes a set
of co-occuring attributes and \(z_{n, l=1,\ldots,L}\) denotes the
compressed representation of observation \(x_n\). The likelihood for
Boolean Matrix Factorisation then takes the form $ p(x_{nd}|.) = \sigma[\lambda \tilde{x}_{nd} (1-2 \prod_{l}(1-z_{nl}u_{ld}) ) ]\;, $
where $\sigma$ is the logistics sigmoid and $\tilde{x} = 2x{-}1$. Note that the product over \(l\) is the Boolean disjunction. This model has a global noise
parameter, $\lambda \in \mathbb{R}^+$, that governs the random
flipping of bits. However, due to the heterogeneity in the data, we require a more expressive model
that can capture heteroscedastic noise. To this end, we propose the
\emph{MaxMachine}. Here, the noise for each latent dimension is
governed by a separate parameter and each data point is generated from
the least noisy associated latent dimension. Hence the model retains
composability and is easily interpretable.
The likelihood takes the form
$  p(x_{nd}|.) = \sigma\left[\tilde{x}_{nd} \max_l (\lambda_l z_{nl} u_{ld})\right],$
where $\lambda_l \in \mathbb{R}$.
Using the max-operation, the latent dimensions compete for
explaining the observations. The \emph{winner} is 
the most accurate predictor and gets to fully explain the
observation. In order for the model to be well defined we have an
additional, clamped latent dimension with \(u_{l,d}{=}z_{nl}{=}1 \;\forall\;(n,d)\).
We propose a beta prior on each $\sigma(\lambda_l)$ and binomial priors on the cardinality of the rows of $U$. The latter can encode our prior belief that the number of co-occuring attributes in each set is much smaller than the total number of attributes.\\
Now, we include the product-type information by adding another layer of
matrix factorisation in the spirit of a \textit{Bayesian hierarchical model}.
This means that the prior on the matrix of latent representation is
factorised according to another MaxMachine moandel. Here, the
higher-order object specific factor matrix is fixed to an encoding of the product type
in a one-hot fashion.

\subsection{Inference}
\label{sec:orge79dcef}
The inference task amounts to estimating, both, the attribute sets
and their assignments and is combinatorially challenging.
We develop sampling based posterior inference, an approach that
has been shown to outperform competing methods in Boolean Matrix
Factorisation\space{}\cite{Rukat2017}. Computation of the full
conditional probabilities of each variable \(u_{ld}\) or \(z_{nl}\)
generally depends on the variables full Markov blanket. We make use
of several
algorithmic tricks and leverage the purely binary states of all
variables as well as the lack of interaction between dimensions that is
induced by the max operation. This enables efficient updates, such that
the algorithm converges for hundreds of thousands of data points
within few minutes on a laptop. After every sweep through all entries of $U$ and $Z$, we set
all $\lambda_l$ to their MAP estimate which is analytically available.
Following posterior inference, we can compute Monte Carlo estimates
of the posterior predictive and thus predict applicability.




\section{Experiments and Results}
We consider a snapshot of the Amazon German marketplace that have been found to be particularly important to customers.
In order to simplify analysis and evaluation we stratify the data into
19 different clusters corresponding to related types of products, such
as \texttt{clothing}, \texttt{computers} or \texttt{jewellery}. Each
cluster consists of products from a variety of product types (1-50). In particular products from different clusters share only very few attributes and therefore share no co-occurrence patterns of mutual relevance. We train on only 500 randomly sampled products from each product type, since a further increase in training data has no effect on the quality of the results. This is due to the redundancy in the binary data.\\
We measure test-set performance by treating randomly selected
product-attribute pairs as unobserved during training and evaluate the area under the ROC curve for the posterior predictive on these test data points.
For the following, experiments we choose \texttt{binomial(0.1)}
priors on the cardinality of each latent set of 
attributes, reflecting our prior belief that the number of
attributes in each co-occuring set is relatively small. For the 
noise parameters (mapped to $[0,1]$), we use a \texttt{beta(10,1)} prior, encoding our believe 
that attribute sets that are applied to a product are relatively
likely to be actually present in the data.
Based on random search, we choose
the remaining hyperparameters.

\begin{figure}[htbp]
\centering
\includegraphics[width=12cm]{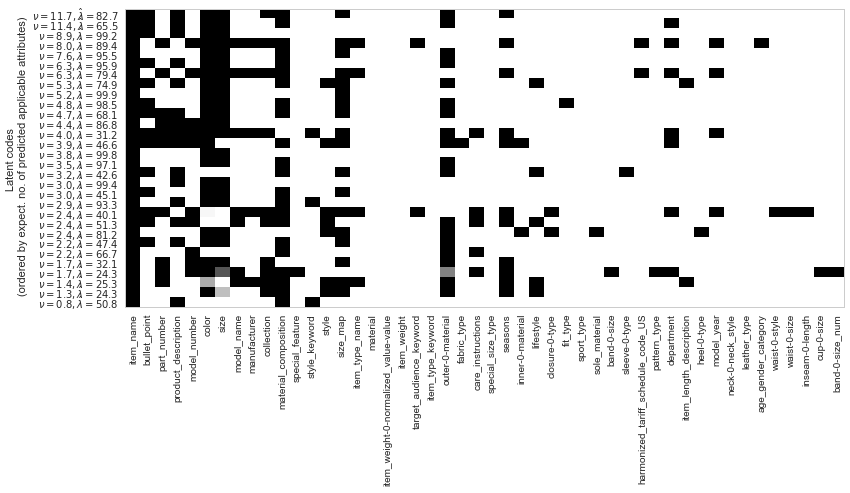}
\caption{\label{fig:orgca0a434}
Patterns of attribute co-occurrence. Shown are posterior means of the
inferred codes \(U\) (black: 1, white: 0). Each row denotes a set of
frequently co-occurring attributes, each column denote an attribute; $\nu$ denotes the expected
percentage of ones in the data that are explained by the corresponding
dimension, $\hat\lambda$ is the average posterior MAP of the noise parameter.}
\end{figure}

We show the inferred sets of co-occurring attributes for clothing
products in Fig.\space{}\ref{fig:orgca0a434}. They indicate reasonable co-occurrences such as (\texttt{waist-style}, \texttt{waist-size}, \texttt{inseam-length},~\ldots).
The corresponding posterior predictive achieves a ROC-AUC of 94\% on held out data, while the product type mean reaches close to 93\%. The
moderate improvements in ROC-AUC can partly be attributed to the
fact that 
the product-type mean predictor has a higher certainty for
attributes that are present in almost all products.
Qualitative evaluation shows that the model makes reasonable
predictions, as for instance to add the attribute \texttt{material} to most
products of type \texttt{bra} or the attribute \texttt{manufacturer} to products of type
\texttt{shirt}, more anecdotal evidence is provided in Table
\ref{tab:orgdc754d7}.
We find a
non-zero, but rather low probability of adding the attribute
\texttt{cup-size} to products of type \texttt{bra}. This can be understood by
noting that \texttt{cup-size} occurs for no other product type and,
therefore, cannot be inferred from correlation. While the probability
is low, it is notable evidence if considered in relative terms:
compared to any other product types, products of the type \emph{bra} have 
by far the highest probability of the attribute \emph{cup-size} being
applicable. This suggests the exploration of attribute-specific
thresholds for practical applications.\\
We repeat this experiment across all product types in the catalogues and find that
the  MaxMachine outperforms the baseline
by margins between 1\% and 15\% in 17 out of 19 clusters. 
The slightly weaker performance in the remaining cluster can largely be explained by extremely homogeneous attribute distributions for each of the contained product types.


\begin{table}[htbp]
  \small
\caption{\label{tab:orgdc754d7}\small
Anecdotal evidence -- for three attributes we list the product types that they are most likely applied to by the model. The percentage is the mean probability of being applied for all products in the product type. In brackets we give the mean probability only for those products that do not have corresponding attribute assigned.}
\centering
\begin{tabular}{l|lll|}
Attribute & cup-size & closure-type & leather-type\\
\hline
Product types & Bra \(22 (10) \%\) & Shoes \(48 (18) \%\) & Shoes \(48 (15)\%\)\\
with largest & Swimwear \(3 (2) \%\) & Pants \(24 (10) \%\) & Outerwear \(3 (3)\%\)\\
p(apply) & Underwear \(3 (2) \%\) & Shorts \(6 (3) \%\) & Shorts \(2(2)\%\)\\
 & Shoes \(2 (2) \%\) & Outerwear \(4 (2) \%\) & (< \(1\%\))\\
 & Suit \(2 (2) \%\) & Bra \(2 (2)\%\) & \\
\end{tabular}
\end{table}



\section{Conclusion}
\label{sec:org5d08bba}
We have described the problem of attribute applicability in the
Amazon catalogue and developed a latent variable model for the
denoising of product/applicability matrices.
Due to the lack of ground truth data we have optimised
reconstruction accuracy conditional on a model that describes an
intuitive generative process, resembling the real-world
procedure of a seller, assigning attributes to products.
As a baseline we have used product-type specific patterns of
applicability and improved in the area under the ROC curve on
hold-out data for 17 out of 19 product clusters.
Anecdotal evidence confirms that our model makes reasonable
predictions.\\
In a practical scenario, more prior expert knowledge
might be available and of high importance. Many types of such
information can be flexibly integrated in the proposed
model. For instance the presence of certain important attributes could be clamped
for certain types of products, while it is inferred for others.
For future work, it would be desirable to avoid the subjective
choice of hyperparameters. In particular the use of non-parametric
priors could lead to a more principled choice of the latent
dimensionality and help model convergence.

\section{Acknowledgements}
We thank Felix Biessmann and David Salinas for helpful discussions, their insights and their expertise that greatly assisted this research. 

\small



\end{document}